\documentclass{sistedes}
\usepackage{amsmath,amsfonts,amssymb}

\usepackage{natbib}
\usepackage{listings}
\usepackage{graphicx}
\usepackage{float}
\usepackage{multirow}
\usepackage{subfig}
\usepackage{threeparttable}

\begin{document}


\title{Incorporating Expert Opinion on Observable Quantities into Statistical Models - A General Framework}

\author{
Philip Cooney\inst{1}\orcidID{0000-0001-9649-0195}
\and
Arthur White\inst{2}\orcidID{0000-0002-7268-5163}
}

\institute{
School of Computer Science and Statistics, Trinity College Dublin, Ireland; \\
\email{phcooney@tcd.ie}\\
\and
School of Computer Science and Statistics, Trinity College Dublin, Ireland;\\
\email{arwhite@tcd.ie}\\
}

\maketitle

\small

\begin{abstract}

This article describes an approach to incorporate expert opinion on observable quantities through the use of a loss function which updates a prior belief as opposed to specifying parameters on the priors. Eliciting information on observable quantities allows experts to provide meaningful information on a quantity familiar to them, in contrast to elicitation on model parameters, which may be subject to interactions with other parameters or non-linear transformations before obtaining an observable quantity. Translating the information elicited on the observable space for use within the statistical model typically involves specifying priors and their associated parameters, so that when the transformation from the parameter to the observable space, the inverse distribution function matches the quantities elicited from the expert. These methods are typically model specific and in some cases require cognitively burdensome elicitation exercises. The approach to incorporating expert opinion described in this paper is distinctive in that we do not specify a prior to match an expert's opinion on observed quantity, rather we obtain a posterior by updating the model parameters through a loss function. This loss function contains the observable quantity, expressed a function of the parameters, and is related to the expert's opinion which is typically operationalized as a statistical distribution. Parameters which generate observable quantities which are further from the expert's opinion incur a higher loss, allowing for the model parameters to be estimated based on their fidelity to both the data and expert opinion, with the relative strength determined by the number of observations and precision of the elicited belief. Including expert opinion in this fashion allows for a flexible specification of the opinion and in many situations is straightforward to implement with commonly used probabilistic programming software. We highlight this using three worked examples of varying model complexity including survival models, a multivariate normal distribution and a regression problem.   
\end{abstract}


\keywords{elicitation, observable space, expert opinion, loss function}

\section{Introduction}

In Bayesian analysis there is an explicit allowance for quantitative subjective judgement in the analysis, however, in a majority of the analyses such information is not incorporated \cite{Mikkola.2021}. There are many situations in which expert opinion can be used in statistical analysis, however, it is particularly important in the absence or scarcity of data to inform probability distributions or for informing inputs for mechanistic models \citep{Garthwaite.2005}. Examples include survival analysis where an expert can be asked for the expected survival probability at a certain time-point which is unobserved due to censoring or eliciting a probability for the prediction of the health care utilization \citep{OHagan.2019} with other applications from fields such as metrology, agriculture, economics and finance detailed by \cite{OHagan.2006}. 

Given the frequency of situations in which data is unavailable but sensible assumptions are required, it is logical to suppose that expert opinion would be formally included decision problems utilizing statistical models, however, that is clearly not the case. As noted by \cite{Kadane.1998}, expertise in a subject-matter is not the same as expertise in statistics and probability, therefore, elicitation of the required inputs can involve training of the experts in statistical concepts and multiple workshops to gain agreement between the experts on a particular input. Consequently, the consensus from the literature it is more beneficial to query the expert about model observables than model parameters \citep{Kadane.1998,Garthwaite.2005, Mikkola.2021}. The underlying elicitation space is the observable space and the form of elicitation can be referred to as ``indirect'' elicitation. The model observables are variables (e.g. model outcomes) that can be observed and directly measured, in contrast to latent variables (e.g. model parameters) that only exist within the context of the model and are not directly observed \citep{Mikkola.2021}.

The literature detailing methods for including opinion on observable quantities primarily falls into two distinct categories (with a comprehensive review by \cite{Mikkola.2021}). The first approach assessing considers the prior predictive distribution $$\pi(x)= \int L(x|\theta)p(\theta)d\theta,$$ and typically asks experts to elicit quantiles of the data (with $p(\theta)$ the prior and $L(x|\theta)$ the likelihood). Using these quantiles the hyperparameters are then derived so that the prior predictive distribution matches these opinions \citep{Percy.2004}. In more complex statistical models hyperparameters required for degrees of freedom parameters and covariance matrices are elicited using other strategies such as imagining hypothetical data and evaluation of quantiles based on conditional distributions \citep{Kadane.1980,Al-Awadhi.1998}.

The second approach focuses on eliciting opinion on the expected value of a response or quantile (conditional on covariates) and the expert's uncertainty about their estimate, akin to how the uncertainty about a parameter estimate is related through the standard error. This information is then used to define the prior on the model parameters (i.e. $\boldsymbol{\beta}$ coefficients) with \cite{Bedrick.1996} providing details for generalized linear models (GLMs) and \cite{Johnson.1996} for survival data modelled with a log-normal distribution. \cite{Hosack.2017} provide a method for including expert opinions on the expected response of GLMs by assuming the $\boldsymbol{\beta}$ coefficients from the linear component are multivariate normal ($\mathcal{MNV}$) and minimizing the Kullback-Leibler divergence between the quantiles of the expected response from the $\mathcal{MNV}(\beta)$ (transformed using the appropriate link function) and the expert's quantiles. An advantage of this approach is that it is model agnostic, so that one method is used to induce priors for different types of GLMs. This is in contrast to \cite{Bedrick.1996} who require model specific calculations to derive the induced prior.     

An alternative approach, which is the basis of this paper, is an approach by \cite{Bissiri.2016} in which expert belief can be incorporated through the use of a loss function. A valid and coherent update of a prior $p(\theta)$ to the posterior $\pi(\theta|x)$ is through a (negative) exponentiated loss function (consistent with the notation of \cite{Bissiri.2016}) is

\begin{equation}\label{eq:BHW-equation}
\pi(\theta|x) \propto \exp\{-l(\theta,x)\}p(\theta).
\end{equation}

Importantly this is just the standard Bayesian update if the loss function is the ``negative log-likelihood'' $l(\theta,x) = -log\{L(x|\theta)\}$. The key idea in this paper is that the loss function will typically include parameters of a probability distribution $\boldsymbol{\phi}$, describing the expert’s opinion about the observable quantity which itself is a function of the model parameters $g(\theta)$ (rather than \cite{Bissiri.2016} who considered opinion on the parameter space), therefore we replace $x$ with $\boldsymbol{\phi}$ and $\theta$ with $g(\theta)$ so that the loss function is

\begin{equation}\label{eq:BHW-equation2}
\pi(\theta|\boldsymbol{\phi}) \propto \exp\{-l(g(\theta),\boldsymbol{\phi})\}p(\theta).
\end{equation}

A key point to highlight is that with this approach (and distinct to the others described above) is that the focus is not on the prior and finding hyperparameters which describe the expert's opinion on the observable space, but rather a posterior distribution which encodes the expert's beliefs about the observable space. Although this posterior can be used as a prior for further update with data we naturally would include the likelihood of the data in Equation \ref{eq:BHW-equation2} for a posterior including both expert opinion and data.

As we will show through examples in the subsequent sections, this approach allows us to be very flexible with respect to the types of information we wish to include in the statistical analysis. Importantly it is very straightforward to include the loss function within existing Bayesian software such as \cite{Stan.2020} with $\text{target }+= -l(g(\theta),\boldsymbol{\phi})$ being used to increment the log probability (code for all the examples discussed in this paper available in the Supplementary Material)\footnote{In NIMBLE it is very straightforward to define a density function to represent the negative loss function, while in JAGS it can be included using the zeros trick.}. In Section \ref{section:Expo-Examples}  we describe incorporating expert opinion with an exponential model, comparing two previous approaches which consider this problem, one using the prior predictive method and the other by eliciting opinions on the expected value of the response. We abbreviate the approach described above as LAP (Loss Adjusted Posteriors) which highlights that a particular prior is updated with a loss function to give a posterior which includes the expert opinion. The focus of the paper is on the construction of the appropriate loss function and the potential influence that priors for the parameters have on the posterior characterizing the expert's opinion.

In Section \ref{section:Elicitation-MNV} we describe the elicitation of a multivariate distribution by using the prior predictive approach to define priors for some of the parameters with expert opinion on the other parameters defined by a loss function. The strategy we employ has much fewer and less complex elicitation questions than previous approaches. In the Section \ref{section:repeated-regression} we describe the inclusion of an expert's opinion on mean change from baseline of a treatment in a longitudinal study. To our knowledge inclusion of expert opinion within models analysing repeated measurements has not been considered previously and furthermore can easily be extended to standard GLMs.
\newpage
\section{Introductory Illustrative Example - Exponential Likelihood}\label{section:Expo-Examples}

\subsection{Prior Predictive Approach}
\cite{Percy.2004} considers asking experts to specify tertiles by stating two times, $Q(1/3)$ and $Q(2/3)$, such that the lifetime of an object was equally likely to be in each of these three intervals: less than $Q(1/3)$; between $Q(1/3)$ and $Q(2/3)$; greater than $Q(2/3)$. Assuming a gamma prior and an exponential sampling distribution, the prior predictive distribution is know as the Lomax distribution (with cumulative distribution function $F(x) = 1-(\frac{\beta}{x+\beta})^\alpha$) and it is straightforward to find the parameters which satisfy the values of $Q(1/3)$ and $Q(2/3)$.

\begin{figure}[ht]
\centering
\scalebox{0.5}{\includegraphics[scale=1]{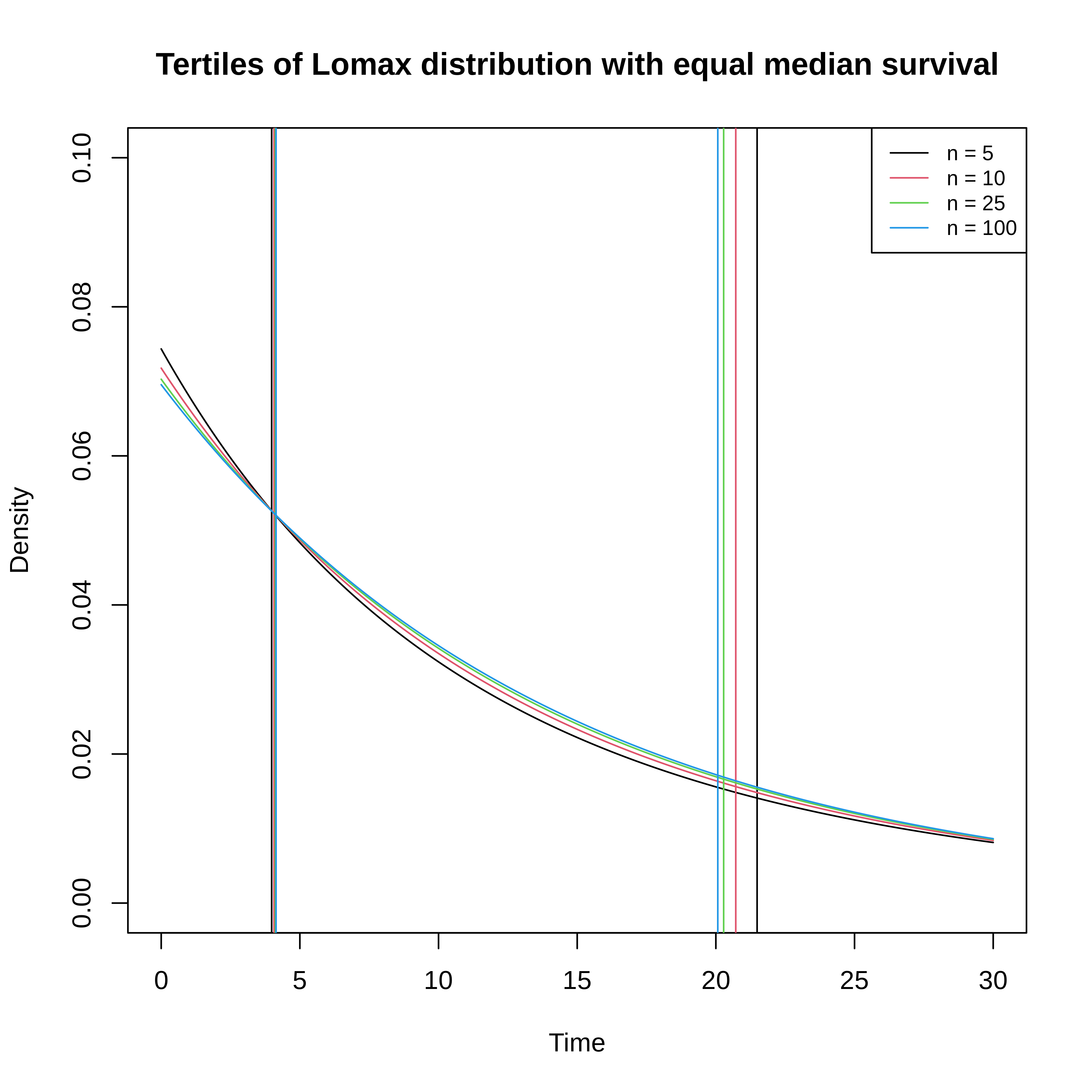} }
\caption{Quantiles of Lomax distributions representing different levels of certainty.}
\label{fig:Percy-Prior}
\end{figure}

It is challenging to encode an expert's uncertainty into the analysis using this approach. Figure \ref{fig:Percy-Prior} shows the $Q(1/3)$ and $Q(2/3)$ tertiles of four different gamma distributions which have the same median survival but different levels of effective samples sizes (ESS), i.e., $\alpha = n = 1, 10, 25 \mbox{ and } 100$. It may not be clear to the expert why their opinion becomes more or less informative by changing the values of $Q(1/3)$ and $Q(2/3)$, especially as small changes in the percentiles can result in very large changes in the informativeness of the prior. One potential solution could be to elicit only one percentile and also elicit an effective sample size, which in this situation is the $\alpha$ parameter.

It is also challenging to apply this approach when considering alternative sampling distributions to the exponential. \cite{Percy.2004} discuss but do not implement an approach for the Weibull distribution, for which they note there is no analytic expression for the prior predictive distribution. This approach then requires numerical methods for both the evaluation of the prior predictive distribution and identification of the hyperparameters, of which four are required. 

\subsection{Expert Opinion on Expected Response and Quantiles}

\cite{Bedrick.1996} discuss using a class of priors termed data augmentation priors (DAP) to include expert opinion on observable outcomes. DAP have the same form as the likelihood, with the idea that the prior for the parameters is based on ``prior observations'' that give rise to a likelihood that has the same form as the likelihood for the data. This results in a posterior that is also in the same form as the likelihood. \cite{Bedrick.1996} discuss DAP for an exponential likelihood including expert beliefs at different values of a covariate, however, for the purpose of comparison we will specify data augmentation priors for the ``intercept-only'' model. For this model the DAP is constructed as $\psi \sim \mathcal{IG}(\alpha,\alpha\tilde{y})$ (inverse gamma) where $\alpha$ is the sample size and $\tilde{y}$ is a parameter representing the mean survival time and $\psi$ represents a random draw of mean survival\footnote{The rate parameter $\lambda = \frac{1}{\psi}$ is the more common parametrization for the exponential model.}. In this scenario we could ask the expert their certainty (as a probability) about the proportion of people surviving past a certain timepoint.

Letting $\tau_i$ be the probability that survival exceeds a certain value $\gamma_i$ at a particular timepoint $t_i$,  we have the following equalities:
$\tau_i = \text{Pr}(\exp\{-t_i/\psi\}> \gamma_i) = \text{Pr}\left(\frac{\alpha\tilde{y}}{\psi} < \frac{-\alpha\tilde{y} \log(\gamma_i)}{t_i}\right)$
where $\frac{\alpha\tilde{y}}{\psi} \sim \mathcal{G}(\alpha,1)$ and $\mathcal{G}$ refers to a gamma distribution.

\begin{figure}[ht]
\centering
\scalebox{0.5}{\includegraphics[scale=1]{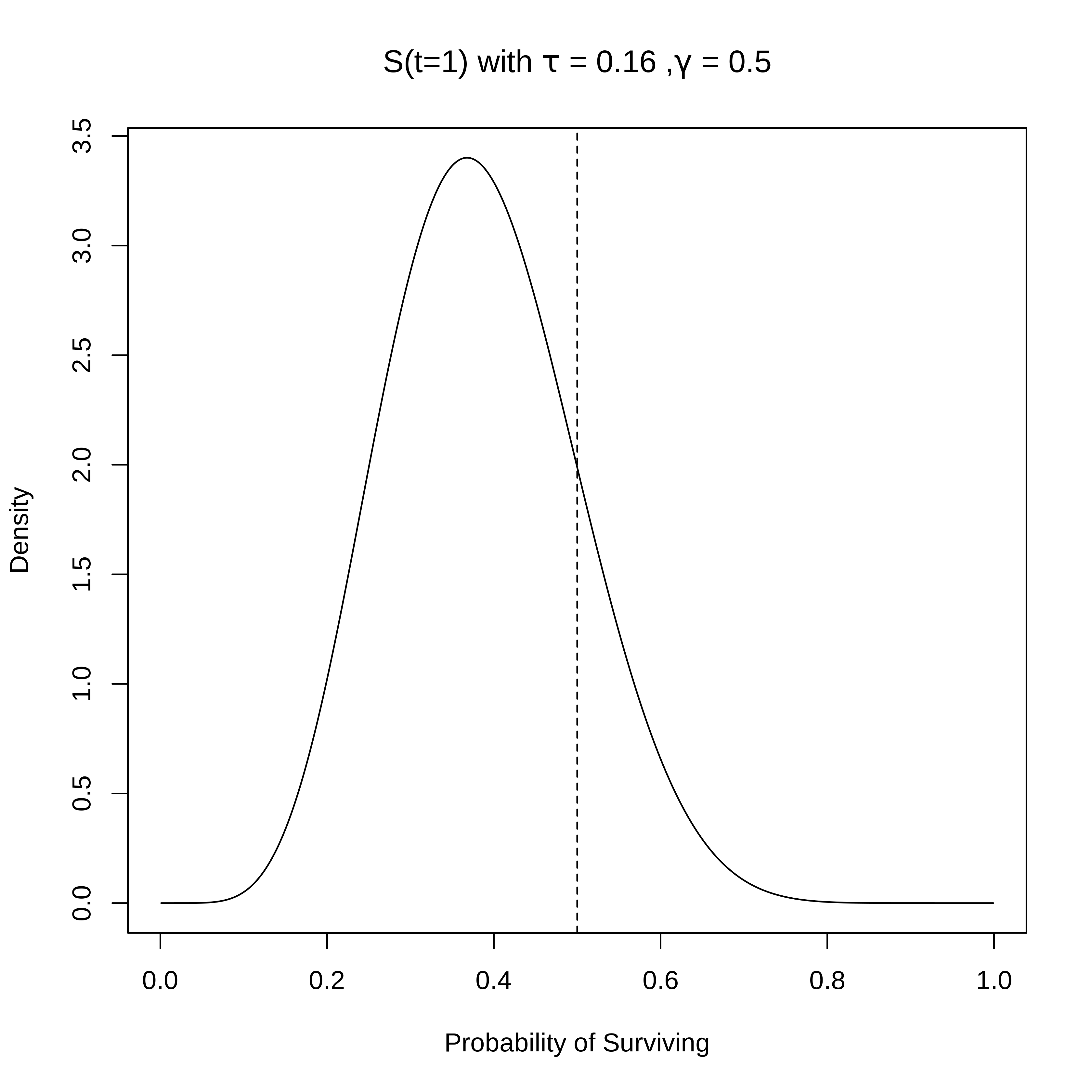} }
\caption{Prior on Survival at t = 1 based on $\mathcal{IG}(10,10)$ prior}
\label{fig:Bedrick-Prior}
\end{figure}

If we elicit $\tau_i$ and $\gamma_i$ for two different timepoints we can derive $\tilde{y}$ and $\alpha$. We could also specify $\alpha$ as the ESS and only elicit one timepoint (and consequently suppress the subscript $i$.) For example, assuming we have an inverse gamma distribution with $\alpha = 10$ and $\tilde{y} = 1$, the probability of survival exceeding $\gamma$ at time $t = 1$ is $ \tau \approx 0.16$ as shown in Figure \ref{fig:Bedrick-Prior}.

\begin{figure}[ht]
\centering
\scalebox{0.5}{\includegraphics[scale=1]{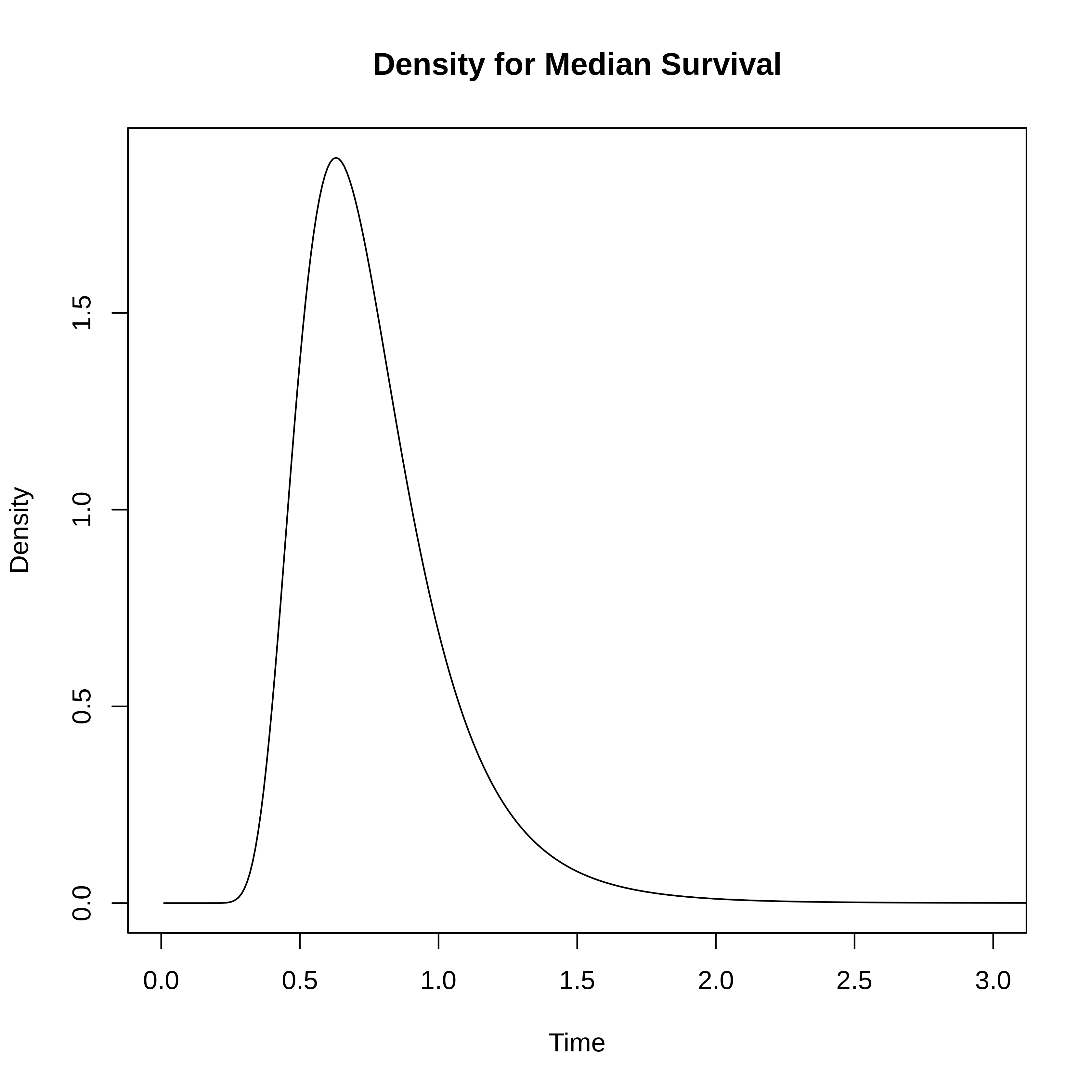} }
\caption{Prior on Median Survival based on $\mathcal{IG}(10,10)$ prior}
\label{fig:Bedrick-Prior-Median}
\end{figure}

We can specify percentile(s) for the median survival ($t_{\text{med}}=\log(2)\psi$) and, after the appropriate calculations for the change of variables,  relate opinion about median survival to the inverse gamma distribution. In Figure \ref{fig:Bedrick-Prior-Median} we see the resulting expert belief on median survival after they have specified the 50\% percentile ($Q(1/2)$) for the median is $\approx 0.71$ with an effective sample size of $\alpha = 10$.

\subsection{Loss Adjusted Posteriors}\label{section:BHW-Approach}

One attractive property of incorporating expert opinion using DAP is that they clearly show the contribution of the expert's opinion in terms of observations, i.e. the ESS of a prior, however, these priors are limited to a number of special cases as described by \cite{Bedrick.1996,Johnson.1996} and deriving the hyperparameters involves non-trivial calculations. By specifying the hyperparameters of the DAP based on expert elicited quantiles, we define a particular density on the observable quantity, such as Equation \ref{eq:DAP-dens} for the median survival (where for clarity $\beta = \alpha\tilde{y}$). 

\begin{equation}\label{eq:DAP-dens}
  f(t_{\text{med}}|\alpha, \beta) =  \frac{\beta^\alpha}{\Gamma(\alpha)}\bigg(\frac{t_{\text{med}}}{log(2)}\bigg)^{-\alpha-1}\exp\bigg(\frac{-\beta log(2)}{t_{\text{med}}}\bigg)\frac{1}{log(2)}.  
\end{equation}

In some situations, however, the expert may wish to not only provide the quantiles but may also wish to specify a lepokurtic distribution to provide a degree of robustness to their opinion or possibly specify a non-parametric histogram prior \citep{OHagan.2006}. 

Finally, and most importantly the approach should be as general as possible or model agnostic, in that it is straightforward to specify expert opinion on different observable for different types of models. One approach which satisfies these requirements is to incorporate information on the model observables through a loss function.  One clear advantage is that the expert is not restricted in the distributions that they can use to describe their belief about the observable quantities. For illustration we assume that the expert would like assume that their belief about median survival is log-normally distributed and this is straightforward when incorporating as a loss function. For the purpose of illustration we consider a log-normal distribution with the same mean and variance as that presented in Figure \ref{fig:Bedrick-Prior-Median}, which through method of moments has the parameters $\mu_{\text{expert}} = -0.32$ and $\sigma_{\text{expert}} = 0.34$.

To compute the loss function we express the model parameters in terms of the observable quantity which for the median survival and exponential likelihood is $g(\psi)=t_{\text{med}}=\log(2)\psi$. According to the expert's belief this opinion is log-normally distributed $\mathcal{LN}(\mu_{\text{expert}}, \sigma_{\text{expert}})$ and the loss function comprises of the deviation of  $t_{\text{med}}$  generated by $\psi$  to the expert's belief. Therefore the loss function $l(g(\psi)|\mu_{\text{expert}}, \sigma_{\text{expert}})$ includes this contribution as $-\log(\mathcal{LN}(\log(2)\psi|\mu_{\text{expert}}, \sigma_{\text{expert}}))$\footnote{It should be noted that the negative will cancel in Equation \ref{eq:BHW-equation2} and it is more simply stated as incrementing the log probability by $\log(\mathcal{LN}(\log(2)/\psi|\mu_{\text{expert}}, \sigma_{\text{expert}}))$, however, we write it as such to be consistent with \cite{Bissiri.2016}.}. If we defined the loss contribution as above and used uniform prior for $\psi$, the Markov Chain Monte Carlo (MCMC) posterior distribution for $\psi$ would produce a median survival which will match the expert's opinion for $t_{\text{med}}$ $\mathbf{\text{exactly}}$ (i.e. a $\mathcal{LN}$ distribution), however, if we parameterized in terms of $\lambda$ and placed a uniform prior on $\lambda$, the posterior distribution for $t_{\text{med}}$ will $\mathbf{\text{not}}$ be exactly $\mathcal{LN}$, as a uniform prior on $\lambda$ will not equate to a uniform prior on the median survival and therefore contribute information to $t_{\text{med}}$ in addition to the expert's opinion. 

To see this clearly, consider consider the transformation from $\psi$ to $t_{\text{med}}$. By change of variables, a uniform prior on $\psi$ will yield a uniform prior on $t_{\text{med}}$ and the contribution of this prior to Equation \ref{eq:BHW-equation2} is constant and will not attenuate the information implied by the expert. Next consider the one-to-one relationship between $\lambda = \log(2)/t_{\text{med}}$ and calculate the density on median survival that a particular prior distribution for $\lambda$ implies. Setting a $\mathcal{U}$(a = 0.001, b = 10) (uniform) prior for $\lambda$ and by change of variables the prior density of the median survival is $p(t_{\text{med}}) = \frac{1}{b-a} \frac{log(2)}{t_{\text{med}}^2} = \frac{1}{b-a} \frac{\lambda^2}{\log(2)}$ (with $a$ and $b$ the upper an lower bounds of $\lambda$). 

Because have we have a closed form expression we can include it explicitly in the loss function, noting that we are cancelling out the prior contribution of the uniform prior for $\lambda$ on the median survival and because we take the negative of the loss function it is equivalent to adding the log of this density to the loss function. 
Therefore the final expression for the loss function is in terms of $\lambda$ is: 
\begin{equation}\label{eq:loss-func}
    -\log(\mathcal{LN}(\log(2)/\lambda|\mu_{\text{expert}}, \sigma_{\text{expert}})) + \log\bigg(\frac{1}{b -a} \frac{\lambda^2}{\log(2)}\bigg).
\end{equation}

Using this loss function to update a uniform prior on $\lambda$ will give a posterior distribution for $\lambda$ which which has a log-normal median survival i.e. the LAP. In most multivariate examples it will not be possible to derive a closed form expression of the density of the observable outcome implied by the prior for the parameters. In order to deal with such situations we can typically reparameterize the model so that the observable quantity is a parameter with a prior distribution (typically uniform) and one of the model parameters is a function of the observable quantity (and the other model parameters). In this hierarchical model specification, the model parameter is a logical or deterministic function observable quantity (and the other model parameters). In the exponential example we specify $t_{\text{med}}$ with a $\mathcal{U}$(a = 0.001, b = 10) prior and because this is a constant we do not require it in the loss function (2nd element in Equation \ref{eq:loss-func}).  

\begin{figure}[H]
\centering
\scalebox{0.5}{\includegraphics[scale=1]{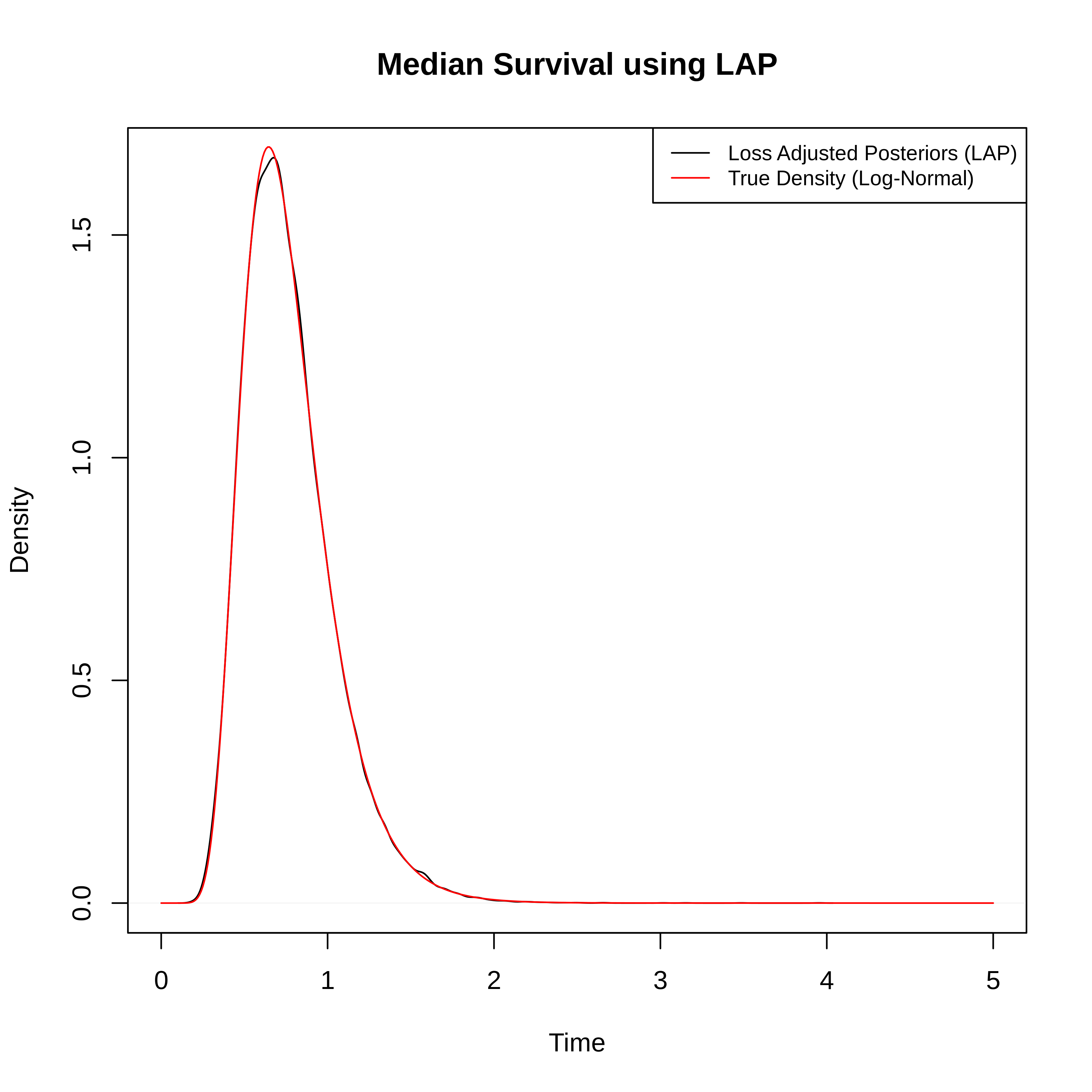} }
\caption{Posterior of Median Survival induced by loss function}
\label{fig:BHW-approach-LN}
\end{figure}

The real strength of this approach is the simplicity at which it can be implemented with psuedocode for Stan model provided below which produces Figure \ref{fig:BHW-approach-LN}:

\begin{lstlisting}[language=R]
 transformed parameters{
 //lambda is a deterministic function of median St 
   lambda = log(2)/median_St;
 }
 model{
  median_St ~ uniform(0.001, 10);
  target += lognormal_lpdf(median_St|mean_expert, sd_expert)
 }
  \end{lstlisting}

Another strength is that we could even implement the expert's belief non-parametrically using the histogram method or parameterize the expert opinion as a (truncated) Student's t distribution with a low number of degrees of freedom. As with any elicitation exercise (either on the observable or parameter space) it is worth quantifying the ESS of the information. The ESS for an exponential likelihood under a gamma or inverse-gamma prior for $\lambda$ or $\psi$ parameters respectively is known through the conjugacy of the one-parameter exponential families (although the ESS will differ by 1 depending on the parametrization). Therefore it is most convenient to fit a gamma distribution to the quantiles of a posterior distribution for $\lambda$ (i.e. using the \texttt{SHELF} package \cite{SHELF}) with the ESS being the shape parameter.  
 
\newpage
\section{Complex Elicitation Problems - Multivariate Normal distribution and Regression Problem}\label{section:MNV-Example}
In this section we describe the incorporation of expert opinion for two more complex statistical models to highlight the flexibility of the approach. In the case of the multivariate normal model we provide an approach which improves upon existing methods in that it requires fewer and less complex queries during the elicitation exercise. In the second example we consider a repeated measures regression problem, a class of problems which to our knowledge has not been analysed with expert opinion on the observable space but under the approach described in the paper can be accomplished with a few lines of extra code.  

\subsection{Expert Opinion on a Multivariate Normal Distribution}\label{section:Elicitation-MNV}

\textbf{Overview of previous approaches}

The problem of including expert opinion with the multivariate normal sampling model has been explored using a natural conjugate prior (normal inverse-Wishart) and a non-conjugate prior (normal generalized inverse-Wishart) \cite{Al-Awadhi.1998, Garthwaite.2001}. The natural conjugate prior forces a dependence between the mean and the covariance, so \cite{Garthwaite.2001} proposed assessment tasks that allow the expert to quantify separately assessments about each of these parameters. 

 In both approaches, assessment tasks include conditional and unconditional quantiles where the conditions were specified by hypothetical data. For example, the degrees of freedom parameters (for the multivariate-t distribution - the prior predictive distribution for the multivariate normal) is assessed by considering the magnitude of difference between two random samples and assessing the median of this absolute deviation for each component $Z_i$. Then, the experts are asked to suppose that two more observations are sampled from the population for which the magnitude of difference is calculated $Z_{i}^*$. These hypothetical values must not be what the expert was ``expecting'' (i.e. $Z_i$) and the expert is then to assesses their conditional median of $Z_{i}^*$, with the ratio of these quantities used in the calculation of the degrees of freedom. The idea is that if the expert's conditional distribution $Z_{i}^*$ is changed only by a small amount relative to $Z_i$, then they have a strong belief about the spread of the multivariate distribution. As noted by \cite{Daneshkhah.2010} ``these are difficult assessments for the expert to make, as it is hard to judge how to change one’s beliefs in light of hypothetical data, particularly as this necessarily has to be done without writing down a prior distribution and applying Bayes’ theorem''. Furthermore, the method requires a substantial number of quantities to be elicited, for a multivariate data with a dimension equal 4 they asked the expert for 50 quantities which is cognitively burdensome. 
\newpage
\textbf{Overview of proposed approach}\\

In the approach described below we can elicit the parameter of the multivariate normal distribution based on $k(k-1)/2 +2k +1$ elicitation's where $k$ is the dimension of the distribution (so in this case 15). Firstly we wish to elicit the expert's strength of belief and we do so by asking them to imagine the number of observations that their opinion represents, clarifying to them that one observation is a random sample of dimension $k$ (this question can also be left for the end of the elicitation exercise) which we denote as $n_e$.

The model priors are as follows:
\begin{align*} 
\Sigma &\sim  \text{LkjCorr}(\eta) \\
 (\mu, \tau) &\sim  \mathcal{NG}(\mu_0,\gamma, \alpha, \beta) \text{ for } 1\dots k
\end{align*}
where $\Sigma$ is a \emph{correlation} rather than covariance matrix, meaning that this prior does not influence the variance components (unlike the Wishart distribution or it's inverse parametrization) and $\mathcal{NG}$ represents a NormalGamma distribution. 

\begin{figure}[H]
\centering
\scalebox{0.5}{\includegraphics[scale=1]{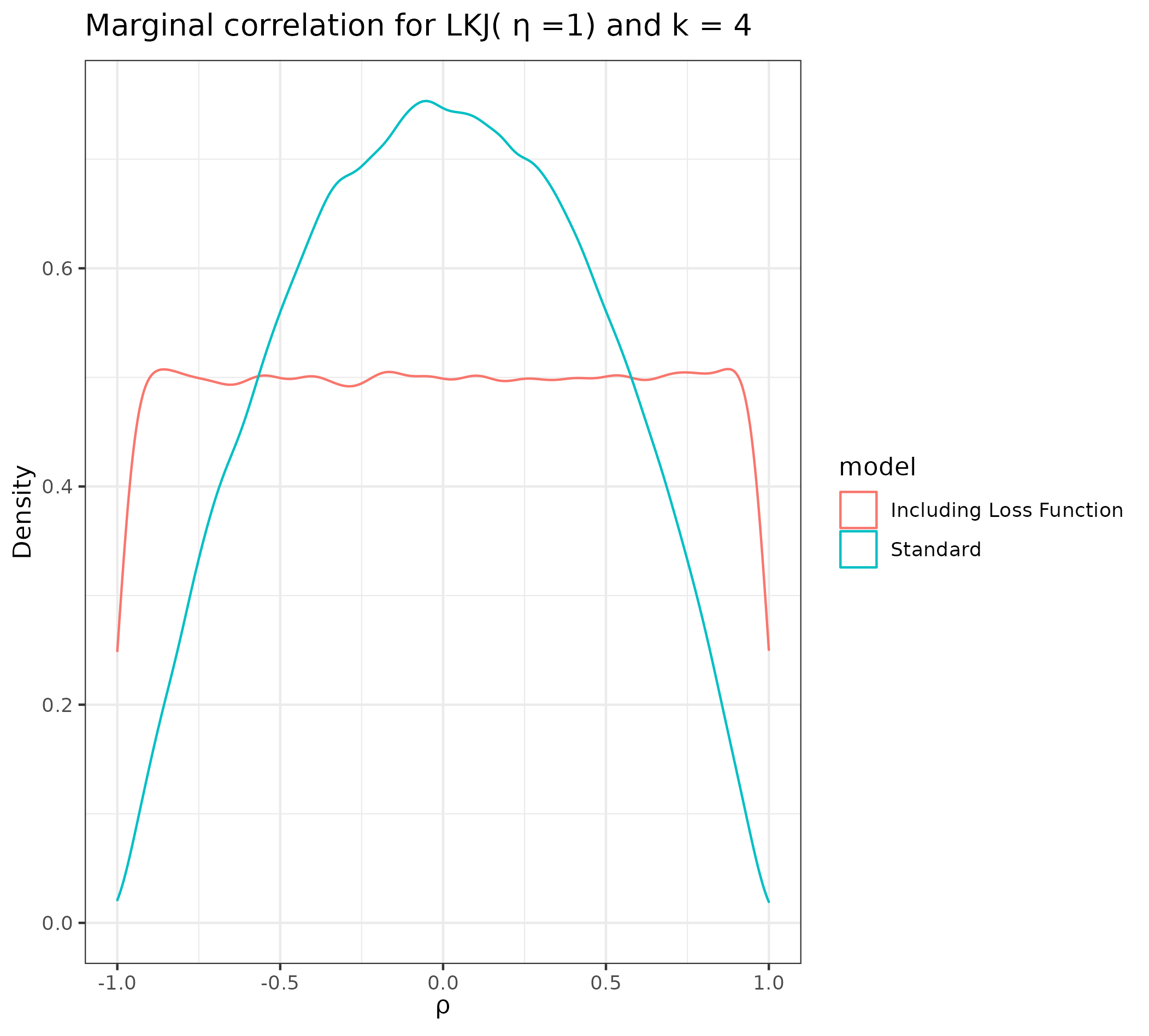} }
\caption{Comparison of densities when including of loss function vs. the standard specification of LKJ distribution}
\label{fig:Prior-Median-Comparison}
\end{figure}

We can remove the impact of the prior by including specifying the loss function as the log density of the LKJ prior which makes it uniform over it's support $[-1,1]$. Although a straightforward observation, it is nevertheless non-trivial because with increasing $k$ the marginal prior probability for the partial correlations become non-uniform and more concentrated around 0. Because the marginal distribution of the partial correlation is proportional to a $\mathcal{B}(\eta-1 + k/2,\eta-1 + k/2)$ (beta), it is possible to obtain a uniform marginal by setting $\eta = (4-k)/2$ so that we obtain a $\mathcal{B}(1,1)$ distribution, however, clearly this is not possible for $k\geq 4$ as $\eta \leq 0$.  Figure \ref{fig:Prior-Median-Comparison} presents the marginal distribution for the LKJ prior with $\eta =1, k = 4$ compared to the density of LKJ distribution including the loss function. 

For each of the $k$ components we have the mean $\mu$ and the precision (inverse variance) $\tau$ from the $\mathcal{NG}$ distribution. The conditional distribution of $\mu$ given $\tau$ is $\mathcal{N}(\mu_0,\gamma\tau)$ with the Normal distribution parameterized in terms of the precision, while $\tau$ is sampled from a $\mathcal{G}(\alpha,\beta)$ distribution. The hyperparameters $\mu_0$ represents the mean value of $\mu$\footnote{Technically more accurate to say median as mean will not exist for $\alpha \leq 0.5$} while $\gamma$ is a scale factor which can be interpreted as the number of pseduo-observations ($n_e$) for the $\mu$ parameter and similarly the precision is estimated from $2\alpha$ pseudo observations so that $\alpha = n_e/2$.   
The prior predictive distribution for a $\mathcal{NG}$ distribution is a (non-standard or scaled and shifted Student's-t) $\mathcal{S}t(\mu_0,\frac{\beta(\gamma+1)}{\alpha*\gamma},2\alpha)$\footnote{for clarity $\mathcal{S}t(x|\mu,\zeta,\alpha) = c\bigg[1+ \frac{\zeta}{\alpha}(x-\mu)^2\bigg]^{-(\alpha +1)/2}$ where $c = \frac{\Gamma(\alpha +1/2)}{\Gamma(\alpha/2)\Gamma(0.5)}(\frac{\zeta}{\alpha})^{1/2}$} where $2\alpha$ is the degrees of freedom of the distribution \cite{Bernardo.2000}. Because we have independence between the correlation matrix and other parameters each component within the $k$ dimension vector has the Student's-t distribution specified as above.

We proceed to identify the hyperparameters of the $\mathcal{NG}$ distribution. For each component in $1\dots k$ we elicit two quantiles of the prior predictive distribution e.g. 0.5 and 0.75. With these parameters along with $n_e$ we can find the parameters of the t-distribution described above by minimizing the squared error. 

We now turn our attention towards the elicitation of the partial correlations $r_{ij}$. \cite{Fackler.1991} considers asking the experts for a concordance probability:
$p_{ij} = \text{P}(\theta_i > \mu_i, \theta_j > \mu_j) \text{ or } \text{P}(\theta_i < \mu_i, \theta_j < \mu_j) $,
i.e. the probability that both $\theta_i$ and $\theta_j$ are either above their expected values or below their expected values. For the bivariate case this probability is $p_{ij} = 0.5 + \sin^{-1}(r_{ij})$\footnote{Owing to the properties of the multivariate normal distribution we can simply drop the elements not relating to $i$ or $j$ to obtain the bivariate distribution. For a simple derivation of the formula see \cite{Kepner.1989} which holds for general multivariate distribution (i.e. one not centered on zero with variance equal 1). It may be worth highlighting that the prior probability of the data is from a $\mathcal{S}t$ distribution, and although no formulas exist for this concordance probability we have verified by simulation that they are the same.} with $r_{ij}$ being the product moment correlation we require for the correlation matrix. 

For each combination of the $k$ variables we elicit the median concordance probability $\tilde{p_{ij}}$ (as it is invariant to transformations). Assuming that the expert is equally confident in their beliefs about the concordance values as they are about the quantiles elicited above (though this is not a requirement) we use $n_e$ to derive the uncertainty in the estimate of $p_{ij}$. To do this by first transforming $\tilde{p_{ij}}$ to $\tilde{r_{ij}}$ and then using Fisher's transformation $F(r) =\text{artanh}(r)$ (the inverse hyperbolic tangent function), noting that this has a normal distribution with mean $\text{artanh}(\tilde{r_{ij}})$ and standard error $\psi=\frac{1}{\sqrt{n_e-3}}$ \cite{Fisher.1915}. Therefore the loss function also includes $-\log(\mathcal{N}(\text{artanh}(\rho_{ij})|\text{artanh}(\tilde{r_{ij}}),\psi))$ summed across across all possible partial correlations (i.e. 6 for $k = 4$) where $\rho_{ij}$ is the partial correlation associated with the correlation matrix generated from $\text{LkjCorr}(\eta)$.  

To summarize, the loss function includes a component which expresses the deviation of the (transformed) partial correlations from the Fisher's transformation of the partial correlation derived from the median concordance probability $p_{ij}$ with the implied uncertainty calculated from the expert's sample size $n_e$. The second component in the loss function is the log-density of the prior on the correlation matrix to obtain uniform prior over the correlations ($\text{LkjCorr}(\eta)$). We have also used the prior predictive approach to estimate the hyperparameters for the marginal (normal) distributions which to our knowledge has not been done before and shows how LAP can integrate expert opinion incorporated on the parameter's priors with expert opinion for other parameters included using a loss function.
\newpage
\textbf{Expert Opinion applied to Multivariate Normal Model}

In this example we consider an example of the method applied to a imagined elicitation exercise. We assume that the dimension of the multivariate normal data is $k = 4$. The expert has been asked to assess their effective sample size which they believe to be $n_e = 10$. For each of the four marginal distributions, they specify the 0.5 and 0.75 quantiles as in Table \ref{tab:Expert-Quantile}.  

\begin{table}[H]
\centering
\caption{\label{tab:Expert-Quantile} Quantiles supplied by the expert and associated hyperparameters.}
  \begin{threeparttable}
    \begin{tabular}{rrrrr}
& \multicolumn{2}{c}{Quantile} & \multicolumn{2}{c}{Hyperparameters}\\
  \hline
 k & 0.5 & 0.75 & $\mu_0$ & $\beta$ \\ 
  \hline
1 & 5.00 & 6.35   & 5 &  16.89\\ 
  2 & 2.00 & 2.67 & 2 & 4.22   \\ 
  3 & 1.00 & 1.34 & 1 & 1.06 \\ 
  4 & 3.00 & 5.02 & 3 & 38\\ 
   \hline
\end{tabular}
   \begin{tablenotes}
      \small
      \item $\gamma$ and $\alpha$ will be equal to $n_e$ and $n_e/2$ by definition.
    \end{tablenotes}
    \end{threeparttable}
\end{table}

\begin{figure}[H]
  \centering
  \subfloat[Concordance Probability]{\includegraphics[width=0.5\textwidth]{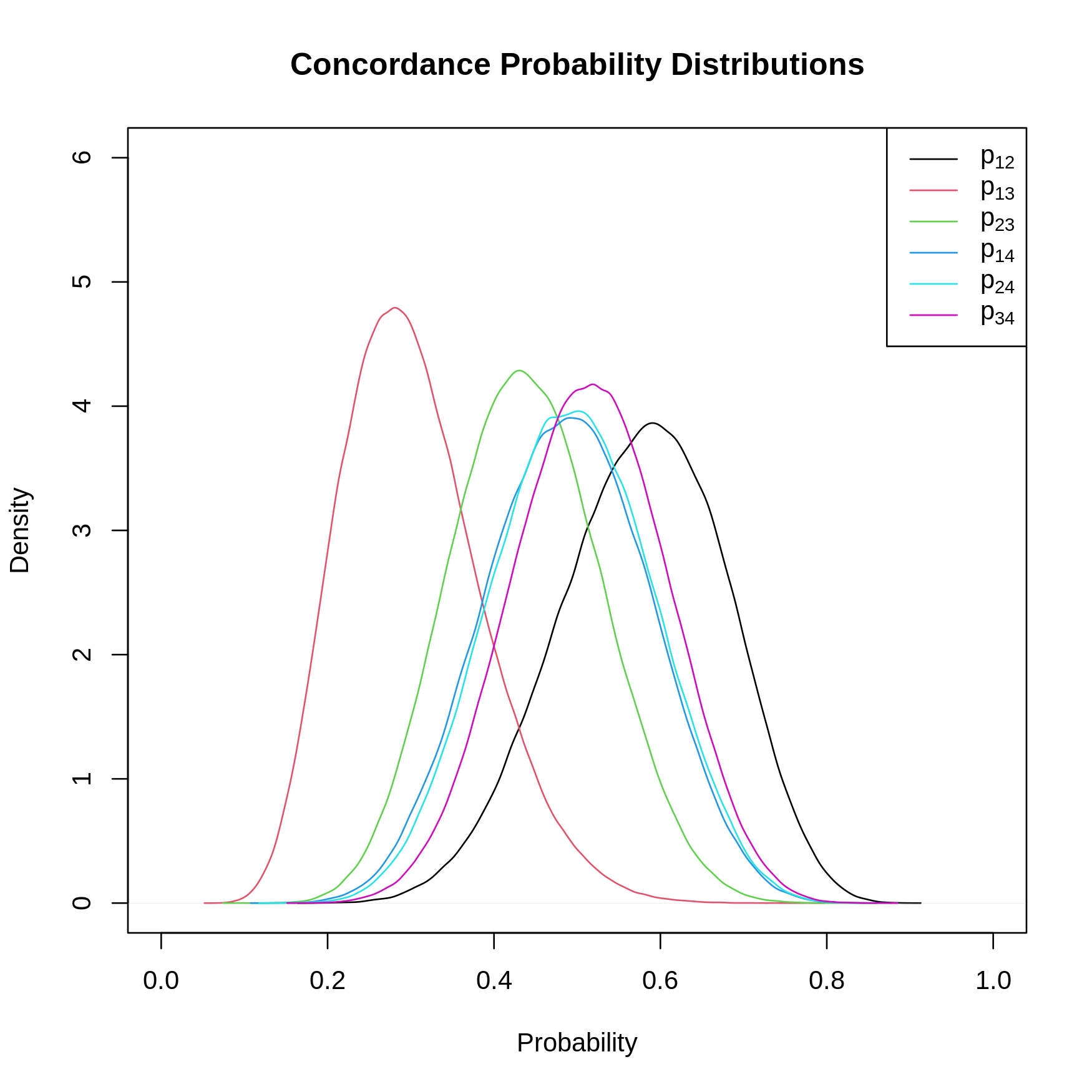}\label{fig:f1}}
  \hfill
  \subfloat[Partial Correlation]{\includegraphics[width=0.5\textwidth]{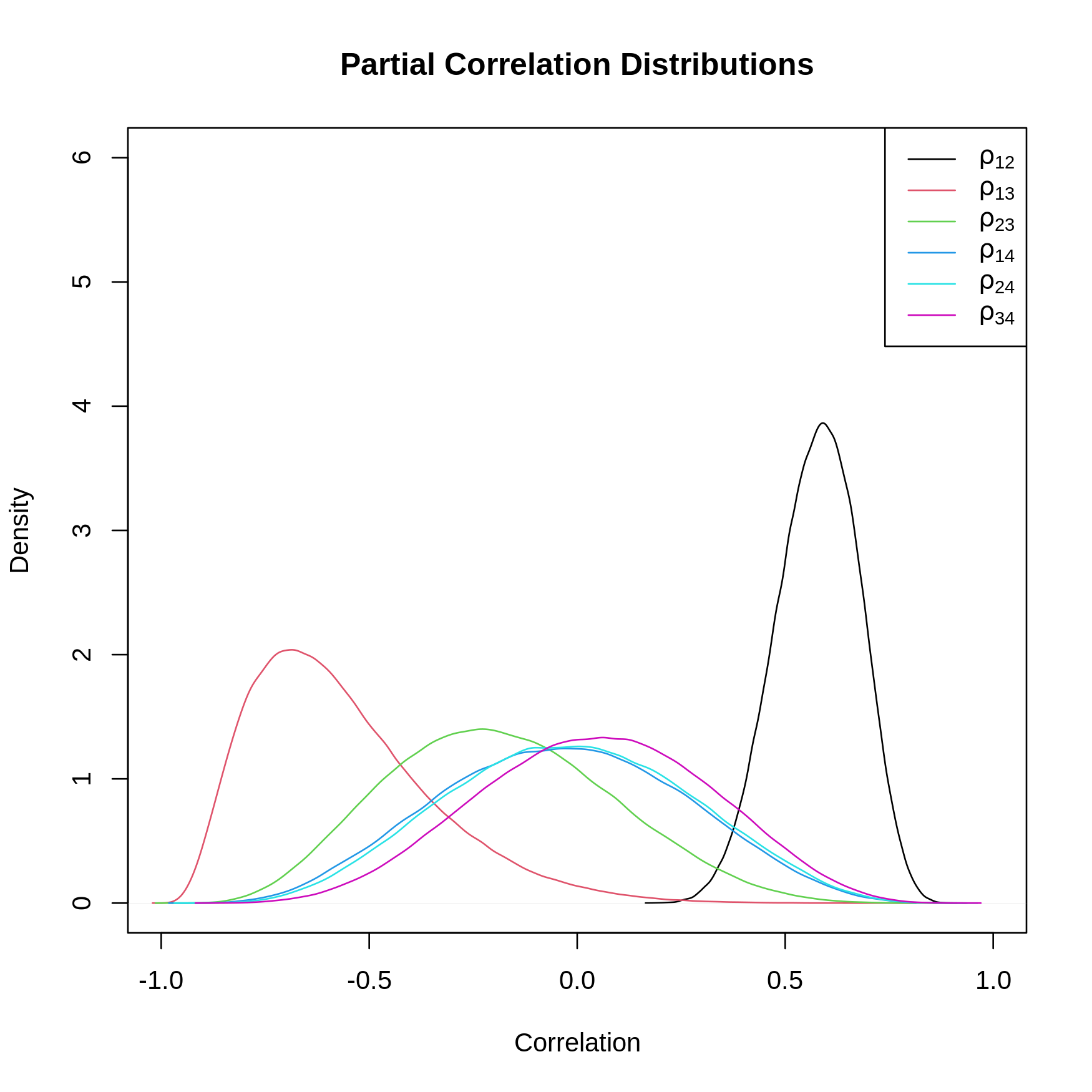}\label{fig:f2}}
  \caption{\label{fig:Expert-Concord-Corr-Plot} Concordance and Correlation induced by loss function }
\end{figure}

For each of the six partial correlations, the expert provides their median concordance probabilities as in Table \ref{tab:Expert-Concord}. Also shown is the median posterior correlations estimated using the loss function. It is worth noting that the median concordance probability for $\tilde{p_{13}}$ from the model is higher than that specified by the expert. The reason for this is that when conditional on the other partial correlations, the partial correlation of the remaining one is restricted to be within a certain interval so that the correlation matrix is semi-positive definite.  We suggest that once the median concordance probabilities are elicited, that the intervals for each concordance probability (conditional on the other concordance probabilities) which produces a positive definite correlation matrix are presented to the expert and situations in which there is substantial density outside the interval the expert is asked to reassess a particular value for coherency. Furthermore, the expert can be shown the distributions of the concordance probabilities (and the correlation parameters) incorporating the loss function (Figure \ref{fig:Expert-Concord-Corr-Plot}) and confirm that it is a reasonable representation of their beliefs.   

\begin{table}[H]
\centering
\caption{\label{tab:Expert-Concord} Median Concordance Probabilities supplied by the expert and those generated by the model.}
\begin{tabular}{rrr}
  \hline
  & Expert's Concordance Probability & Concordance Probability from model \\ 
  \hline
  $\tilde{p_{12}}$ & 0.60 & 0.58 \\ 
  $\tilde{p_{13}}$ & 0.25 & 0.30\\ 
  $\tilde{p_{23}}$ & 0.40 & 0.44\\ 
  $\tilde{p_{14}}$ & 0.50 & 0.49\\ 
  $\tilde{p_{24}}$ & 0.50 & 0.49\\ 
  $\tilde{p_{34}}$ & 0.50 & 0.51\\ 
   \hline
\end{tabular}
\end{table}

\subsection{Expert Opinion on a regression model with repeated measurements}\label{section:repeated-regression}

In the final example we discuss the incorporation of an expert's opinion regarding the expected change from baseline (noting that the approach could alternatively include information on the expected value of a response variable conditional on the covariate) in repeated measures setting.

We consider data presented in \cite{Littell.1990} in which the effect of three different exercise programs on participant's strength is assessed over 7 different timepoints. 
In the original analysis the population effects\footnote{often also called fixed-effects} were modelled with time as a quadratic function (with various covariance structures) and we consider a situation in which the expert was asked for their belief about the \emph{expected} change from baseline for one of the programs (denoted as WI). 

\begin{figure}[H]
\centering
\scalebox{0.5}{\includegraphics[scale=1]{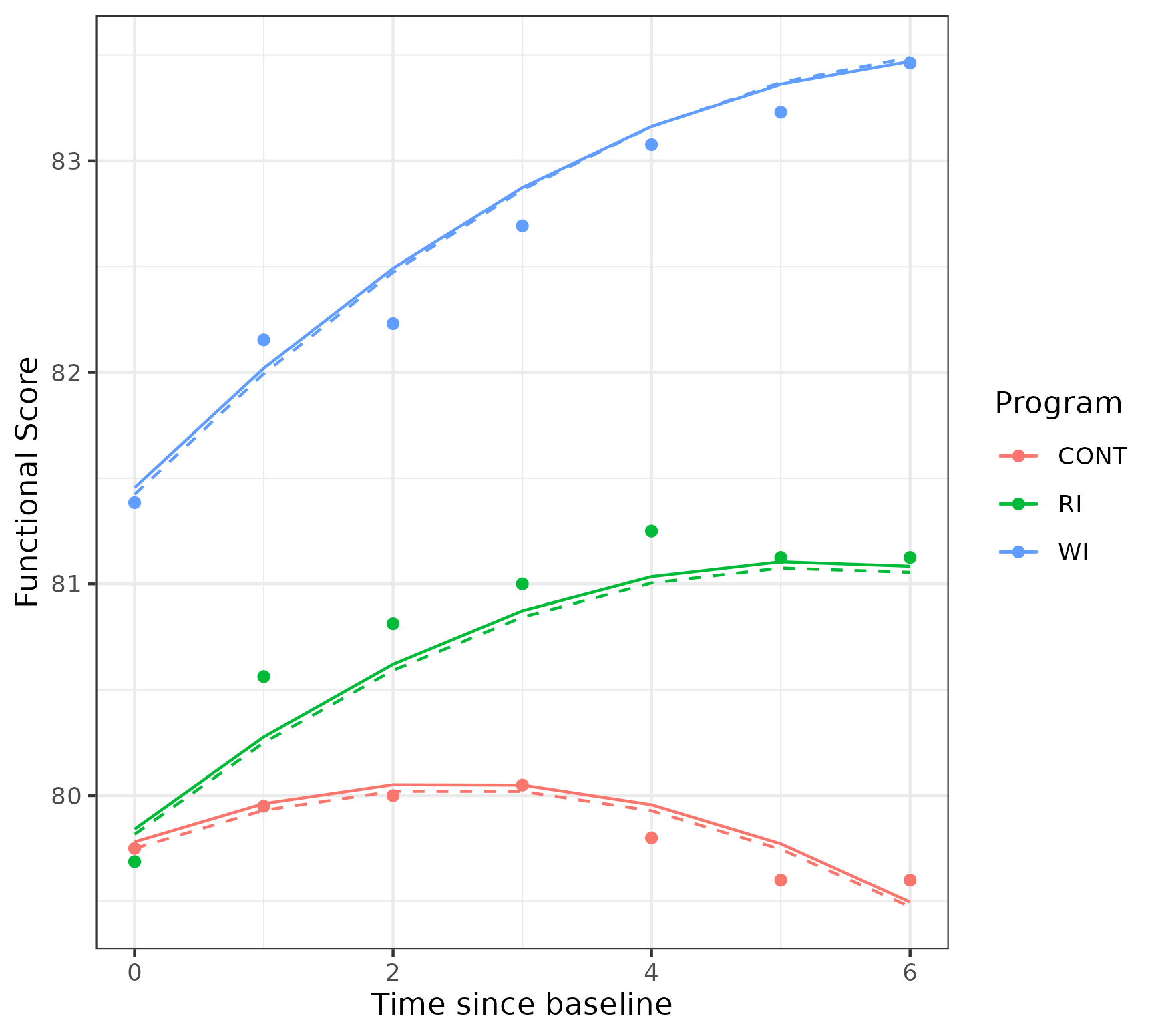}}
\caption{Comparison fixed effect model estimates with and without expert opinion (dashed and full line respectively).}
\label{fig:Repeated-Measures-Model}
\end{figure}

\begin{figure}[H]
\centering
\scalebox{0.5}{\includegraphics[scale=1]{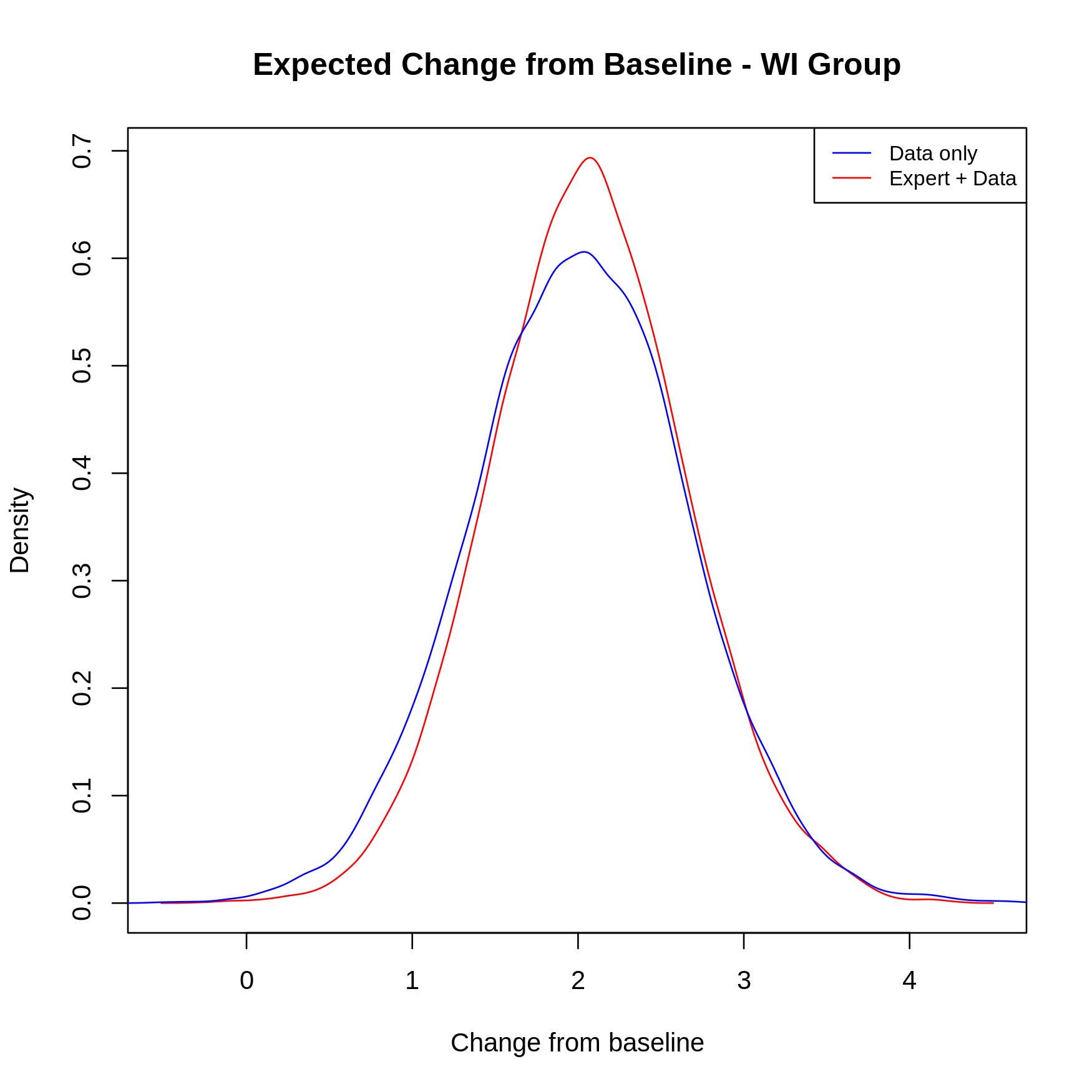}}
\caption{Comparison of mean change from baseline estimates with and without expert opinion (dashed and full line respectively).}
\label{fig:Mu-Diff}
\end{figure}

We assume that the expert beliefs that the mean change from baseline to the final timepoint is a normal distribution with $\mu_{expert} =2.5$ and $\sigma_{expert}=1.5$. Because of the quadratic term it is not possible to place a prior on the population level $\beta$ coefficients to induce this opinion, however, it is straightforward using the loss function in which we simply calculate the difference in the expected value at the final and first timepoints for the WI group, denoted as $\xi$ which is then included in the loss function as $-\log(\mathcal{N}(\xi| \mu_{\text{expert}},\sigma_{\text{expert}}^2))$ (now parameterized in terms of the variance).
In terms of model specification we include the population level effects with time modelled as a quadratic and also include a coefficient for treatment as per \cite{Littell.1990}. To account for the repeated measures we define individual effects for the intercept, slope, and quadratic effect of time\footnote{Also known as random effects}. We specify time (linear and quadratic components) as orthogonal polynomial contrasts to aid estimation. The Stan code was generated using the \texttt{brms} package \cite{brms.2017} which was then modified by the inclusion of the loss function with the impact of the expert's opinion illustrated in Figures \ref{fig:Repeated-Measures-Model} \& \ref{fig:Mu-Diff}. The expert's opinion has very marginally changed the fixed effects regression line, however, the change is more apparent in the posterior distribution for the expected change from baseline. 

We can get a heuristic for the effective sample size of this parametric prior by comparing the standard deviation of the posterior distribution for the change from baseline without expert opinion (0.67) to the standard deviation of the expert's prior (1.5). Considering the value of 0.67 was generated from 13 participants in the WI group the following relation $\sigma = 0.67\sqrt{n_{\text{data}}}\approx1.5\sqrt{n_{\text{expert}}}$ so that $n_{\text{expert}}\approx 2.6$ which seems plausible considering the modest impact of including the data\footnote{This heuristic calculation ignores the potential for the expert opinion to be in conflict with the observed data (as does others such as \cite{Morita.2008,Neuenschwander.2020}); for example if the $\mu_{\text{expert}} = 10$ we would not be correct in stating that the expert opinion is equivalent to 2.6 subjects.}. 

\section{Discussion}
In this paper we describe a general approach to including expert opinions on observable quantities within statistical models. The theoretical justification for the approach is based of \cite{Bissiri.2016}, however, this paper expands upon the idea both in a theoretical sense and by providing a variety of practical examples. \cite{Bissiri.2016} briefly describe an example in which an expert declares that a parameter $\theta$ is close to 0 with a quadratic loss function. The core idea in this paper is that the expert declares that a function of a parameter (which relates to the observable quantity) can be described in some manner such as a probability distribution. This idea which has been applied previously to survival models by \cite{Cooney.2023} and has been extended for a variety of statistical problems in this paper. Additionally, this paper describes the potential for the prior on the parameters to attenuate the information provided by the expert and how to eliminate it's impact. As shown in Section \ref{section:MNV-Example} the specification of a loss function may be of interest in standard data analysis exercises, such as when attempting to model multivariate data in which the correlations are close to $\pm1$, as with higher dimension of the data, the LKJ distribution will place more prior probability on partial correlations close to zero. It is worth highlighting that this approach requires experts to assign subjective probability to deterministic functions of the parameters such as expected values or quantiles and therefore, is in the same category as work by \cite{Bedrick.1996,Hosack.2017}, albeit acknowledging that the focus is not on the prior but a loss function. Within the LAP framework it is not possible to specify beliefs about the prior predictive density as by definition the parameter of interest has been integrated out of the expression (and we cannot specify a deterministic relationship between the data $x$ and the parameter(s) which generated that data). This however does not prevent using the prior predictive approach to define priors which encode expert opinion for some of the parameters with information on other parameters included through the loss function (as in Section \ref{section:MNV-Example}).  

The approach described in this paper addresses many of the key difficulties with respect to expert knowledge elicitation described by \cite{Mikkola.2021}. They cite practical reasons, such as many of the approaches still being too difficult for non-statistical experts to use, and the lack of good open source software that integrates well with the current probabilistic programming tools used for other parts of the modelling workflow. It is natural that eliciting expert's opinions on observable quantities is easier than elicitation on the parameter space, however, it can still be cognitively burdensome. As described in Section \ref{section:Elicitation-MNV} our approach to elicitation of the multivariate normal distribution requires many fewer questions and avoids elicitation about conditional distributions and hypothetical data which are cognitively more challenging. Another key issue that we believe this method addresses is that that integrates well with the current probabilistic programming tools used for other parts of the modelling workflow. Stan and other Bayesian programmes are the default tools for many statisticians and the ease which this approach can be integrated with these tools is highlighted by the pseudo-code in Section \ref{section:BHW-Approach} and that a few lines of code that was required to update the existing code generated by the \texttt{brms} package for the example in Section \ref{section:repeated-regression}.  

Using this approach, it is typically straightforward to include expert opinion on observable quantiles, however, it always worth checking how well the posterior distribution for the expert's opinion (just with the loss function and without data) matches the one elicited from the expert. As noted earlier a uniform prior on the parameters does not necessitate a uniform density on a function of those parameters (i.e. the observable quantity). In many practical situations the choice of vague priors will have a relatively modest impact on the density implied by the expert (which will diminish the more informative the expert's opinion is). Furthermore, in the context of survival analysis \cite{Cooney.2023} found that the posterior distributions with data and loss functions tend to be very similar even with different types of relatively non-informative priors\footnote{They also found very good agreement between the Bayesian method and a frequentist approach motivated as a penalized likelihood.}. In the situation that the impact is non-trivial, it is often possible express the observable quantity (upon which expert opinion was sought) as a parameter, assign a uniform prior to it and express one of the model parameters as a deterministic function of the observable quantity (and the other parameters as required) as described in Section \ref{section:BHW-Approach}. Even if it is not possible to express this relationship analytically, it is usually possible solve this numerically albeit with a large increase in computational burden. 

Finally, although the incorporation of expert opinion with statistical models using this approach is more straightforward, the robustness of the inferences generated from them will rely upon the quality of the information elicited from the experts. \cite{OHagan.2006} provides an in-depth treatment of expert elicitation and it evident that considerable effort (which may take the form of a workshop) is required to obtain methodologically appropriate opinions. Although important to quantify (at least approximately), the informativeness of any type of elicited information included in a statistical model, it is especially important in situations where the expert is providing opinion on the expected value of the response and their uncertainty around that estimate (e.g. Section \ref{section:repeated-regression}) or elicitation's based on prior predictive quantiles. Experts may not understand that uncertainty does not scale linearly with the sample size, or put another way a relatively small reduction in uncertainty can result in a large increase in implied samples size, illustrated in the extreme case in Figure \ref{fig:Percy-Prior}. Therefore in each of our examples we try to produce an approximate estimate of ESS (or as in the case of Section \ref{section:MNV-Example} specify it explicitly) so that the expert can be given the opportunity to revise their estimates at the elicitation stage rather than requiring post hoc adjustments or reassessments. 

\section*{Supplementary Material}
All codes used to generated the figures and results presented in this analysis is available from the following link:\href{https://github.com/Anon19820/loss-codes}{https://github.com/Anon19820/loss-codes}.

\end{document}